\documentclass[a4paper]{report}
\usepackage[utf8]{inputenc}
\usepackage[T1]{fontenc}
\usepackage{RJournal}
\usepackage{booktabs}

\usepackage{natbib}
\usepackage{hyperref}
\usepackage{wrapfig}
\usepackage{amsmath}
\usepackage{caption}
\usepackage{subcaption}
\usepackage[english]{babel}
\usepackage{lscape}
\usepackage{listings}

\begin{document}

\sectionhead{Contributed research article}
\volume{XX}
\volnumber{YY}
\year{20ZZ}
\month{AAAA}

\begin{article}
\newcommand{\fairmodels}{\pkg{fairmodels }}
\newcommand{\fobject}{\code{fairness\_object }}
\newcommand{\fobjects}{\code{fairness\_objects }}
\newcommand{\parityloss}{\textit{parity\_loss }}
\renewcommand{\arraystretch}{1.5}

\title{\pkg{fairmodels}: a Flexible Tool for Bias Detection, Visualization, and Mitigation in Binary Classification Models}
\author{Jakub Wiśniewski, Przemysław Biecek}

\maketitle

\abstract{
Machine learning decision systems are becoming omnipresent in our lives. From dating apps to rating loan seekers, algorithms affect both our well-being and future. Typically, however, these systems are not infallible. Moreover, complex predictive models are eager to learn social biases present in historical data that can lead to increasing discrimination. If we want to create models responsibly then we need tools for in-depth validation of models also from the perspective of potential discrimination.
This article introduces an R package \CRANpkg{fairmodels} that helps to validate fairness and eliminate bias in binary classification models easily and flexibly. The \pkg{fairmodels} package offers a model-agnostic approach to bias detection, visualization, and mitigation. The implemented set of functions and fairness metrics enables model fairness validation from different perspectives. The package includes a series of methods for bias mitigation that aim to diminish the discrimination in the model. 
The package is designed not only to examine a single model but also to facilitate comparisons between multiple models.
}

\section{Introduction}

Responsible machine learning and in particular fairness are gaining attention within a machine learning community. The reason for this is that predictive algorithms are becoming more and more decisive and influential in our lives. This impact could be less or more significant in areas ranging from user's feeds on social platforms, displayed ads, and recommendations at an online store to loan decisions, social scoring, and facial recognition systems used by police and authorities. Sometimes it leads to automated systems that learn some undesired bias preserved in data for some historical reason. Whether seeking a job \citep{8731591} or having one's data processed by court systems \citep{propublica}, sensitive attributes such as sex, race, religion, ethnicity, etc. might play a major role in the decision. Even if such variables are not directly included in the model, they are often captured by proxy variables such as zip code (a proxy for the race and wealth), purchased products (a proxy for gender and age),  eye color (a proxy for ethnicity). As one would expect they can give an unfair advantage to a privileged group. Discrimination takes the form of more favorable predictions or higher accuracy for a privileged group. For example, some popular commercial gender classifiers were found to perform the worst on darker females \citep{pmlr-v81-buolamwini18a}. From now on such unfair and harmful decisions towards people with specific sensitive attributes will be called biased. 

The list of protected attributes may depend on the domain for which the model is built as well as on the country. For example, the European Union law is summarised in the Handbook on European non-discrimination law \cite{European-non-discrimination}, which lists the following protected attributes that cannot be the basis for inferior treatment: sex, gender identity, sexual orientation, disability, age, race, ethnicity, nationality or national origin, religion or belief, social origin, birth, and property, language, political or other opinions. This list, though long, does not include all potentially relevant items, e.g. in the USA, a protected attribute is also pregnancy, the status of a war veteran, or genetic information. 

While there are historical and economical reasons for this to happen, such decisions are unacceptable for ethical reasons and sometimes are prohibited by local law regulations. 
The problem is not simple, especially when the only criterion set for the system is performance. In some problems, we observe a trade-off between accuracy and fairness where lower discrimination, leads to lower performance \citep{kamiran}. Sometimes labels, which are considered ground truth might also be biased \citep{labelwrong} and when controlling for that bias the performance and fairness might improve at the same time. Most of the time however when we want to improve fairness from one perspective it becomes worse in another \citep{barocas-hardt-narayanan}.

The bias in machine learning systems has potentially many different sources. In \cite{mehrabi2019survey} authors categorized bias into its types like historical bias, where unfairness is already embedded into the data reflecting the world, observer bias, sampling bias, ranking and social biases, and many more. That shows how many dangers are potentially hidden in the data itself. Whether one would like to act on it or not, it is essential to detect bias and make well-informed decisions whose consequences could potentially harm many groups of people. Repercussions of such systems can be unpredictable. As argued by \cite{barocas-hardt-narayanan} machine learning systems are even able to aggravate the disparities between groups, which is called by the authors' feedback loops. Sometimes the risk of potential harm resulting from the usage of such systems is high. This was noticed for example by the Council of Europe that wrote the set of guidelines where it states that the usage of facial recognition for the sake of determining a person's sex, age, origin, or even emotions should be mostly prohibited \citep{facialrecognition}.  

Not every difference in treatment is discrimination. \cite{cirillo_sex_2020} presents examples of desirable and undesirable biases based on the medical domain. For example, in the case of cardiovascular diseases, documented medical knowledge indicates that different treatments are more effective for different genders. So different treatment regimens according to medical knowledge are examples of desirable bias. Later in this paper, we present tools to identify differences between groups defined by some protected attribute but note that this does not automatically mean that there is discrimination. 

We would like to also point out that fixing the machine learning model may not be enough in every case, and sometimes the whole design of the data acquisition and/or annotation might cause the model to be biased \citep{barocas-hardt-narayanan}.      

\subsection{Related Work}

Assembling predictive models is nowadays getting easier. Packages like \CRANpkg{h2o} \citep{H2OAutoML} provide AutoML frameworks where non-experts can train quickly accurate models without deep domain knowledge. Model validation should also be that simple.

Two main kinds of fairness are a concern to multiple stakeholders. These are group and individual fairness. The first one concerns groups of people with the same protected attributes (gender, race, etc.). It focuses on measuring if these groups are treated similarly by the model. The second one is focused on the individual. It is most intuitively defined as treating similar individuals similarly \citep{statisticalparity}. Both concepts are sometimes considered to conflict with each other but they don't need to be. If we factor in certain assumptions such as whether the disparities are due to personal choices or unjust structures \citep{reuben}.

Several frameworks have emerged for Python to verify various fairness criteria, the most popular are \textbf{aif360} \citep{aif360-oct-2018}, \textbf{fairlearn} \citep{bird2020fairlearn}, or \textbf{aequitas} \citep{2018aequitas}. They have various features for detection, visualization and mitigation of the bias in machine learning models.
For the R language, until recently the only available tool was the \CRANpkg{fairness} \citep{fairness} package which compares various fairness metrics for specified subgroups. The \pkg{fairness} package is very helpful, but it lacks some features. For example, it does not allow to compare the machine learning models between each other and to aggregate fairness metrics to facilitate the visualization, but most of all it does not give a quick verdict whether a model is fair or not. Package \CRANpkg{fairadapt} aims at removing bias from machine learning models by implementing pre-processing procedure described in \cite{plecko2019fair}. Our package tries to combine the detection and mitigation processes. It encourages the user to experiment with the bias, try different mitigation methods and compare results. The package \fairmodels not only allows for that comparison between models and multiple exposed groups of people, but it gives direct feedback if the model is fair or not (more on that in the next section). Our package also equips the user with a so-called \fobject which is an object aggregating possibly many models, information about data, and fairness metrics. \fobject can later be transformed into many other objects that can facilitate the visualization of metrics and models from different perspectives. If a model does not meet fairness criteria, there are various pre-processing and post-processing bias mitigation algorithms implemented and ready to use. It aims to be a complete tool for dealing with discriminatory models in a group fairness setting. 

In particular, we show how to use this package to address four key questions: How to measure bias? (see section \nameref{bias}), How to detect bias? (see section \nameref{detect}), How to visualize bias? (see section \nameref{visualization}), How to mitigate bias? (see section \nameref{mitigation}).

It is important to remember that fairness is not a binary concept that can be unambiguously defined. The presented tools allow for fairness analysis, thanks to which we will be able to detect differences in the behavior of the model for different protected groups. But such analysis will not guarantee that all possible fairness problems have been detected. 
Like other validation tools, it should be used with caution and awareness.

\section{Measuring and detecting bias}

In model fairness analysis, a distinction is often made between analysis for group fairness and individual fairness. The former is defined by the equality of certain statistics determined on protected subgroups and we focus on this approach in this section. We write more about the latter later in this paper.

\subsection{Fairness metrics}
\label{detect}
Machine learning models just like human-based decisions can be biased against people with certain sensitive attributes which are also called protected groups. They consist of subgroups - people who share the same sensitive attribute, like gender, race, or some other feature. 

To address this problem we need to first introduce fairness criteria. Following \cite{barocas-hardt-narayanan}, we will present these criteria based on the following notation.

\begin{itemize}
    \item Let $A \in \{a,b, ...\}$ mean protected group and values $A \neq a$ denote membership to unprivileged subgroups while $A = a$ membership to privileged subgroup. To simplify the notation we will treat this as a binary variable, but all results hold if $A$ has a larger number of groups.      
    \item Let $Y \in \{0,1\}$ be a binary label (binary target = binary classification) where $1$ is preferred, favorable outcome.
    \item Let $R \in [0,1]$ be a probabilistic response of model, and $\hat{Y} = 1$ when $R \geq 0.5$, otherwise $\hat{Y} = 0$.
\end{itemize}

Table \ref{tab:fairnessTable1} summarises possible situations for the subgroup $A=a$. We can draw up the same table for each of the subgroups.

\begin{table}[h!]
    \centering
    \begin{tabular}{c|cc|c}
        $A = a$ &  $Y = 1$ & $Y = 0$ & \\ \hline
         $\hat{Y} = 1$ & $TP_a = P(Y = 1, \hat{Y} = 1 | A = a)$ & $FP_a = P(Y = 0, \hat{Y} = 1 | A = a)$ & $P(\hat{Y} = 1 | A = a)$ \\
         $\hat{Y} = 0$ & $FN_a = P(Y = 1, \hat{Y} = 0 | A = a)$ & $TN_a = P(Y = 0, \hat{Y} = 0 | A = a)$ & $P(\hat{Y} = 0 | A = a)$ \\ \hline
          & $P(Y = 1| A = a)$ & $P(Y = 0 | A = a)$ \\ 
    \end{tabular}
    \caption{Summary of possible model outcomes for subpopulation $A = a$. We assume that outcome 1 is favourable.}
    \label{tab:fairnessTable1}
\end{table}

According to \cite{barocas-hardt-narayanan} most discrimination criteria can be derived as tests that validate the following probabilistic definitions: 
\begin{itemize}
    \item Independence, i.e. $R \perp A$, 
    \item Separation, i.e. $R \perp A \mid Y$,
    \item Sufficiency, i.e. $ Y \perp A \mid R$.
\end{itemize}
Those criteria and their relaxations might be expressed via different metrics based on confusion matrix for certain subgroup. To check if those fairness criteria are addressed we propose checking 5 metrics among privileged group (a) and unprivileged group (b):

\begin{itemize}
    \item Statistical parity: $P(\hat{Y} = 1 | A = a) = P(\hat{Y} = 1 | A = b)$.
    
    Statistical parity (STP) ensures that fractions of assigned positive labels are the same in subgroups. It is equivalent of Independence \citep{statisticalparity}. In other words, the values in the last column of Table \ref{tab:fairnessTable1} are the same for each subgroup.
    \item Equal opportunity: $P(\hat{Y} = 1 | A = a, Y = 1) = P(\hat{Y} = 1 | A = b, Y = 1)$.
    
     Checks if classifier has equal True Positive Rate (TPR) for each subgroup. In other words, the column normalized values in the second column of Table \ref{tab:fairnessTable1} are the same for each subgroup.
     It is a relaxation of Separation \citep{NIPS2016_6374}.
    \item Predictive parity: $P(Y = 1 | A = a, \hat{Y} = 1) = P(Y = 1 | A = b, \hat{Y} = 1)$
    
    Measures if a model has equal Positive Predictive Value (PPV) for each subgroup. 
    In other words, the row normalized values in the second row of Table \ref{tab:fairnessTable1} are the same for each subgroup.
    It is relaxation of Sufficiency \citep{ppv}.
    \item Predictive equality: $P(\hat{Y} = 1 | A = a, Y = 0) = P(\hat{Y} = 1 | A = b, Y = 0)$.
    
    Warrants that classifiers have equal False Positive Rate (FPR) for each subgroup. 
    In other words, the column normalized values in the third column of Table \ref{tab:fairnessTable1} are the same for each subgroup.
    It is relaxation of Separation \citep{ppe}.
    \item (Overall) Accuracy equality: $P(\hat{Y} = Y | A = a) = P(\hat{Y} = Y | A = b)$.
    
    Makes sure that models have the same Accuracy (ACC) for each subgroup. \citep{accuracy}
\end{itemize}
The reader should note that if the classifier passes Equal opportunity and Predictive equality, then it also passes Equalized Odds \citep{NIPS2016_6374}, which is equivalent to Separation criteria. 

While defining the metrics above we assumed that there are only 2 subgroups. This was done to facilitate notation, but there might be more unprivileged subgroups. A perfectly fair model would pass all criteria for each subgroup \citep{barocas-hardt-narayanan}.

Not all fairness metrics are equally important in all cases. The metrics proposed above are a proposition that aims to give a more holistic view into the fairness of the machine learning model. Practitioners that are informed in the domain may consider only those metrics that are relevant and beneficial from their point of view. For example in \cite{KOZODOI2021} in the fair credit scoring use case authors concluded that the separation is the most suitable non-discrimination criteria. More general instructions can be also found in \cite{EUhandbook} along with examples of protected attributes. Sometimes, however, non-technical solutions to fairness problems might be beneficial. Note that not all types of unfairness will be discovered by group fairness metrics and the end-user should decide whether a model is acceptable in terms of bias or not. 

However tempting it is to think that all the criteria described above can be met at the same time, unfortunately, this is not possible. \cite{barocas-hardt-narayanan} shows that, apart from a few hypothetical situations, no two of $\{\text{Independence}, \text{Separation}, \text{Sufficiency}\}$ can be fulfilled simultaneously. So we are left balancing between the degree of imbalance of the different criteria or deciding to control only one criterion.

Let us illustrate the intuition behind Independence, Separation, and Sufficiency criteria using the well-known example of the COMPAS model for estimating recidivism risk. 
Fulfilling the Independence criterion means that the rate of sentenced prisoners should be equal in each subpopulation. It can be said that such an approach is fair from society's perspective.
Fulfilling the Separation criterion means that the fraction of innocents/guilty sentenced should be equal in subgroups. Such an approach is fair from the prisoner's perspective. If I am innocent, then I should have the same chance of acquittal regardless of sub-population. This was the expectation presented by the ProPublica Foundation in their study.
Meeting the Sufficiency criterion means that among the convicted, there should be an equal fraction of innocents. Similarly, for the non-convicted. This approach is fair from the judge's perspective, If I convicted someone then he should have the same chance of being innocent regardless of the sub-population. This approach is presented by the company developing the COMPAS model, Northpointe.
Unfortunately, as we have already written, it is not possible to meet all these criteria at the same time.

\subsection{Acceptable amount of bias}
\label{bias}
It would be hard for any classifier to maintain the same relations between subgroups. That is why some margins around the perfect agreement are needed. To address this issue, as the default setting we accepted the four-fifths rule \citep{adverseimpact} as the benchmark for discrimination rate which states that \textit{"A selection rate for any race, sex, or ethnic group which is less than four-fifths ($\frac{4}{5}$) (or eighty percent) of the rate for the group with the highest rate will generally be regarded by the Federal enforcement agencies as evidence of adverse impact[...]."} The selection rate is originally represented by statistical parity, but we adopted this rule to define acceptable rates between subgroups for all metrics. There are a few caveats to the preceding citation concerning the size of the sample and the boundary itself. Nevertheless, the four-fifths rule is a good guideline to adhere to. In the implementation, this boundary is represented by $\varepsilon$ and it is adjustable by the user but the default value will be 0.8. 
This rule is often used, but in each specific case, one should see if the fairness criteria should be set differently.

Let $\varepsilon > 0$ be the acceptable amount of a bias. In this article we would say that the model is not discriminatory for a particular metric if the ratio between every unprivileged $b, c, ...$ and privileged subgroup $a$ is within $(\varepsilon, \frac{1}{\varepsilon})$. The common choice for the epsilon is 0.8, which corresponds to the four-fifths rule. For example, for the metric Statistical Parity ($STP$), a model would be $\varepsilon$-non-discriminatory for privileged subgroup $a$ if it satisfies 
\begin{equation}
\forall_{b \in A \setminus \{a\}} \;\;
   \varepsilon < STP_{ratio} = \frac{STP_b}{STP_a} < \frac{1}{\varepsilon}.
   \label{fig:ratio}
\end{equation}

\subsection{Evaluating fairness}

The main function in the \fairmodels package is \texttt{fairness\_check}. It returns  \fobject which can be either visualized or processed by other functions. This will be further explained in the "Structure" section. When calling \texttt{fairness\_check} for the first time the three arguments are: 
\begin{enumerate}
    \item \textbf{explainer} - an object that combines model and data that gives a unified interface for predictions. It is a wrapper over a model created with the \CRANpkg{DALEX} \citep{JMLR:v19:18-416} package.
    \item \textbf{protected} - a factor, vector containing sensitive attributes (protected group) . It does not need to be binary. Each level denotes a distinct subgroup. The most common examples are gender, race, nationality, etc.
    \item \textbf{privileged} - a character/factor denoting level in the protected vector which is suspected to be the most privileged one.
\end{enumerate}

\subsubsection{Example}

In the following example we are using \textit{German Credit Data} dataset \citep{Dua:2019}. In the dataset, there is information about people like age, sex, purpose, credit amount, etc. For each person, there is a risk assessed with taking credit, either good or bad. It will be a target variable. We will train the model on the whole dataset and then measure fairness metrics on it as opposed to training and testing on different subsets, however, this approach is also possible and advisable.  

First, we create a model.

\begin{example}
library("fairmodels")

data("german")

lm_model <- glm(Risk~., data = german, family = binomial(link = "logit"))
\end{example}

Then, create a wrapper that unifies the model interface.

\begin{example}
library("DALEX")

y_numeric <- as.numeric(german$Risk) -1
explainer_lm <- DALEX::explain(lm_model, data = german[,-1], y = y_numeric)
\end{example}

Finally, create and plot the fairness checks.

\begin{example}
fobject <- fairness_check(explainer_lm,
                          protected = german$Sex,
                          privileged = "male")
plot(fobject)
print(fobject)
\end{example}

\begin{figure}[p]
\centering 
\includegraphics[width=0.8\linewidth]{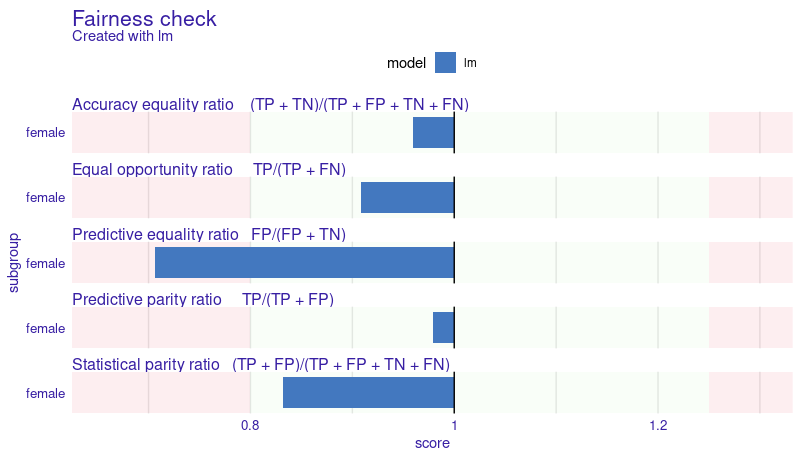} 
\caption{Graphical output for the \texttt{fairness\_check()}. The default value for  $\varepsilon$ is set to 0.8. The light green areas which are values within $(\varepsilon, \frac{1}{\varepsilon})$ signifies an acceptable difference in fairness metrics. Fairness metric names are given along the formulas used to calculate the score in some subgroups to facilitate interpretation. The ratio here means that after metric scores were calculated, the values for unprivileged groups (here female) were divided by values for the privileged subgroup (here men) as in formula (\ref{fig:ratio}).}
\label{fig:fc1}

\vspace{0.5cm}

\includegraphics[width=0.8\linewidth]{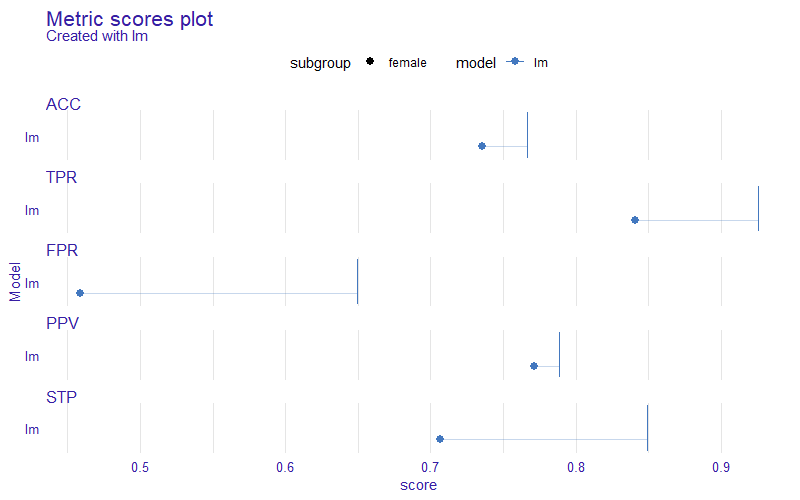} 
\caption{Metric scores plot. It might be helpful for understanding the intuition behind the \texttt{fairness\_check()} from Figure \ref{fig:fc1}. The dots (unprivileged subgroups - female) and vertical lines (privileged subgroup - male) of the metric scores plot represent raw metric scores of subgroups. The horizontal lines act as a visual aid for measuring the difference between the scores of the metrics subgroups. This plot might be a good first point for understanding Figure \ref{fig:fc1}. In fact Figure \ref{fig:fc1} can be directly derived from Figure \ref{fig:fc2_plot}. To do this we need to divide the score denoted by the dot with the score denoted by the vertical line. This way we obtain a value indicated by the height of barplot in \ref{fig:fc1}. The orientation of the barplot depends on whether the value is bigger or lower than 1.  Intuitively the longer the horizontal line in Figure \ref{fig:fc2_plot} (the one connecting the dot with the vertical line) is the longer the bar will be. If the scores of privileged and unprivileged subgroups are the same then the bar will start from 1 and point to one, so will have a height equal to 0.}  
\label{fig:ms_for_fc1}

\vspace{0.5cm}

\includegraphics[width=0.4\linewidth]{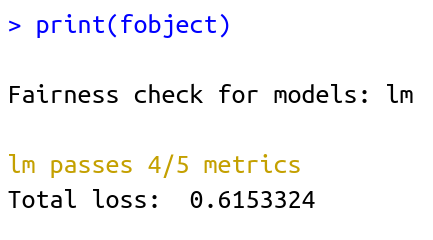} 
\caption{Numerical output for the \texttt{fairness\_check()}. Total loss here should be interpreted as the sum of heights of bars in Figure \ref{fig:fc1}}
\label{fig:fc1_print}

\end{figure}

Figure \ref{fig:fc1} presents the output from the \verb'fairness_check()'. The plot \ref{fig:fc1}  can be obtained by modifying more intuitive plot \ref{fig:ms_for_fc1}. The exact procedure is explained in caption under the figure. 

In this example, fairness criteria are satisfied in all but one metric.  The logistic regression model has a lower false positive rate (FP/(FP+TN))) in the unprivileged group than in the privileged group. It exceeds the acceptable limit set by $\varepsilon$, thus it does not satisfy the Predictive Equality ratio criteria.

For a quick assessment if a model passes fairness criteria \verb'fairness_check()' object might be summarized with the \verb'print()' function as in Figure \ref{fig:fc1_print}. Total loss is the sum of fairness metrics, see equation \ref{eq:parityLoss} for more details.  

It is rare that a model perfectly meets all the fairness criteria. Therefore, a very useful feature is the ability to compare several models on the same scale. In the example below, we add two more explainers to the fairness assessment. Now \fobject (in code: \code{fobject}) wraps three models together with different labels and cutoffs for subgroups. The \fobject can be later used as basis for another \texttt{fairness\_object}. In detail, running \texttt{fairness\_check()} for the first time explainer/explainers have to be provided along with three arguments described at the start of this section. As shown below, when providing explainers with \texttt{fairness\_object}, those arguments are not necessary as they are already part of the object. 

First, let us create two more models based on the \textit{German Credit Data}. The first one will be a logistic regression model that uses fewer columns and has access to \code{Sex} feature. The second is random forest from \CRANpkg{ranger} \citep{ranger}. It will be trained on the whole dataset.

\begin{example}
discriminative_lm_model <- glm(Risk~.,
         data   = german[c("Risk", "Sex","Age", "Checking.account", "Credit.amount")],
         family = binomial(link = "logit"))

library("ranger")
rf_model <- ranger::ranger(Risk ~.,
                           data = german,
                           probability = TRUE,
                           max.depth = 4,
                           seed = 123)
\end{example}

These models differ in the way how the predict function works. To unify operations on these models, we need to create \CRANpkg{DALEX} explainer objects. The \verb'label' argument specifies how these models are named on plots.

\begin{example}
explainer_dlm <- DALEX::explain(discriminative_lm_model,
                                data = german[c("Sex", "Age",
                                                "Checking.account",
                                                "Credit.amount")],
                                y = y_numeric,
                                label = "discriminative_lm") 

explainer_rf <- DALEX::explain(rf_model, 
                                data = german[,-1], 
                                y = y_numeric)
\end{example}

Now we are ready to assess fairness.

\begin{example}

fobject <- fairness_check(explainer_rf, explainer_dlm, fobject)
plot(fobject)
print(fobject)

\end{example}

\begin{figure}[ht]
\centering
    \includegraphics[width=0.7\textwidth]{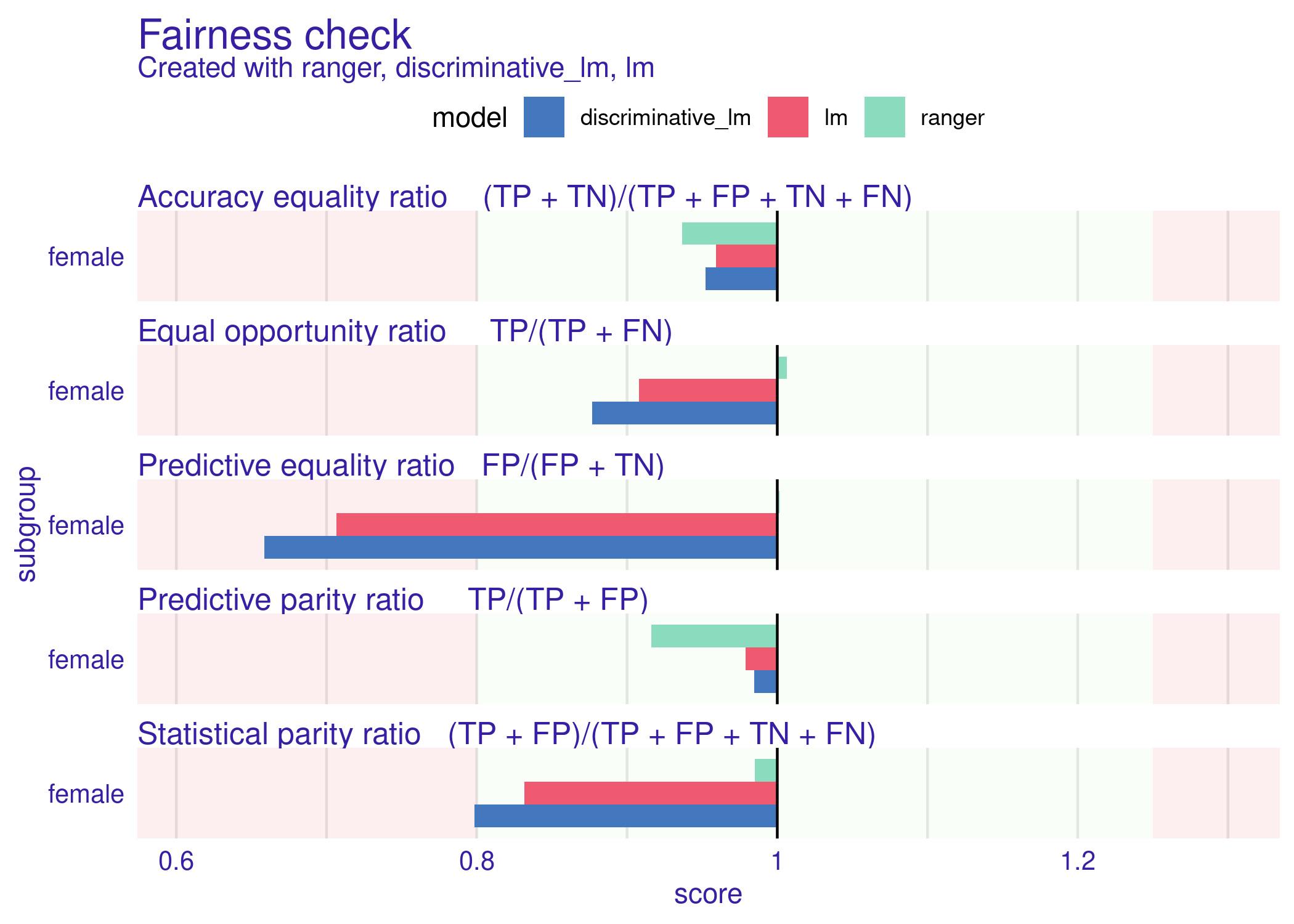}
    \caption{Graphical output for the \texttt{fairness\_check()} for three models. }
    \label{fig:fc2_plot}
\end{figure}

\begin{figure}[ht]
\centering

    \includegraphics[width=0.6\textwidth]{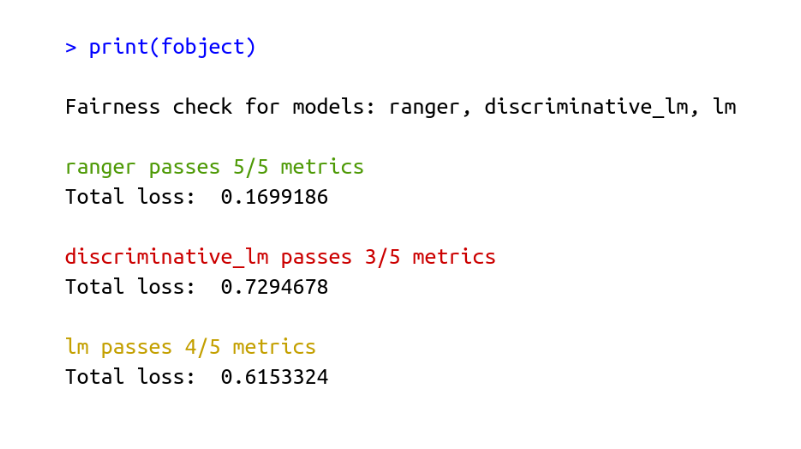}
    \caption{Numerical summary for the \texttt{fairness\_check()} for three models. }
    \label{fig:fc2_print}

\end{figure}

When plotted (Figure \ref{fig:fc2_plot}) new bars appear on familiar plane. Those are new metric scores for added models. Figure \ref{fig:fc2_print} shows the numerical summary for these three models that is printed into the console.

\section{Package Architecture}

The \fairmodels package provides a unified interface for predictive models independently of their internal structure. Using model agnostic approach with \pkg{DALEX} explainer facilitates this process \citep{JMLR:v19:18-416}. For each explainer, there is a unified way to check if explained model lives up to user fairness standards. Checking fairness with \fairmodels is straightforward and can be done with the three-step pipeline. 

\textbf{classification model \%>\% explain() \%>\% fairness\_check() }

The output of such a pipeline is an object of class \fobject which is a unified structure to wrap model explainer or multiple model explainers and other \fobjects in a single container.  Aggregation of fairness measures is done based on groups defined by model labels. This is why model explainers (even those wrapped by \code{fairness\_objects}) must have different labels. Moreover, some visualizations for model comparison assume that all models are created from the same data. Of course, each model can use different variables or use different feature transformations, but the order and amount of rows shall stay the same. To facilitate aggregation of models \fairmodels allows creating \fobjects in other ways: 

\begin{itemize}
    \item explainers \%>\% fairness\_check() - possibly many explainers can be passed to  \code{fairness\_check()}
    \item \fobjects \%>\% fairness\_check() - explainers stored in \fobjects passed to fairness\_check() will be aggregated into one \fobject
    \item explainer \& \fobjects \%>\% fairness\_check() - explainers passed directly and explainers from \fobjects will be aggregated into one \fobject
\end{itemize}

When using the last two pipelines protected vectors and privileged parameters are assumed to be the same, so it is not necessary to pass them to \code{fairness\_check()}

\begin{figure}
\centering
\includegraphics[width=0.9\linewidth]{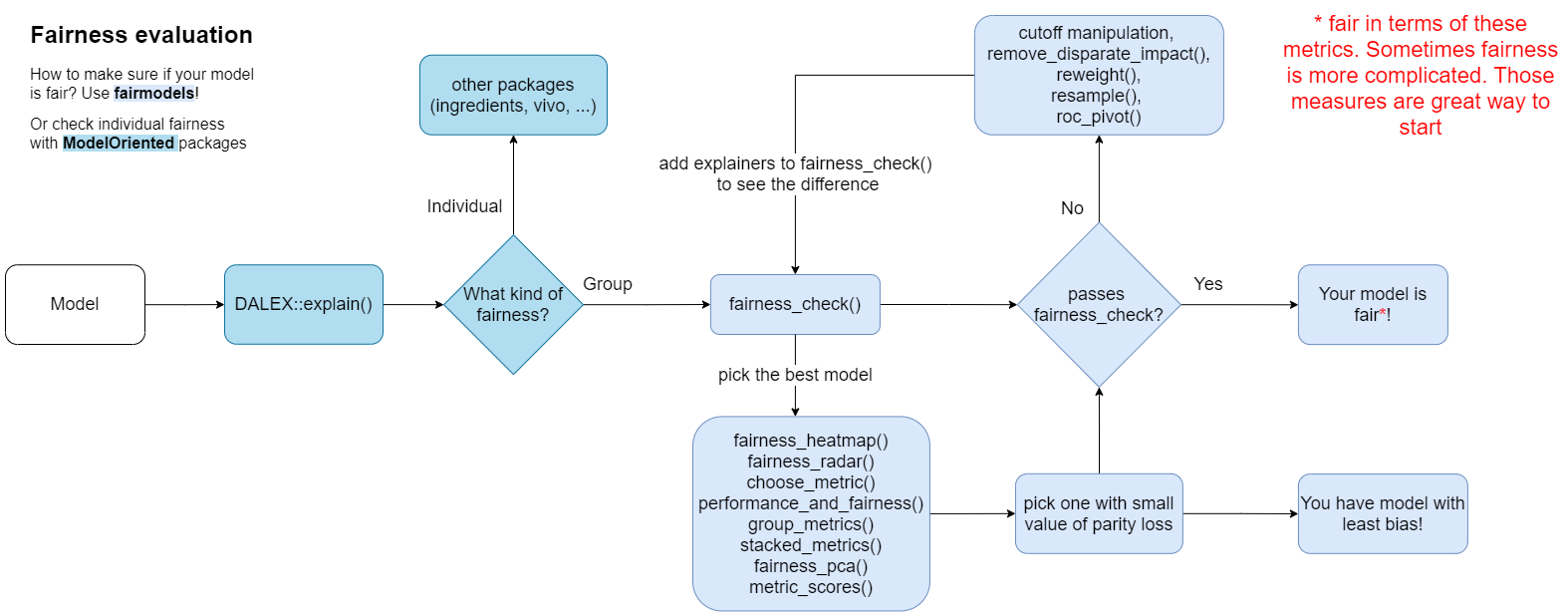} 
\caption{Flowchart for the fairness assessment with the \fairmodels package}
\label{fig:flowchart}
\end{figure}

\begin{landscape}
 \begin{figure}
  \centering
  \includegraphics[width=1.4\textwidth]{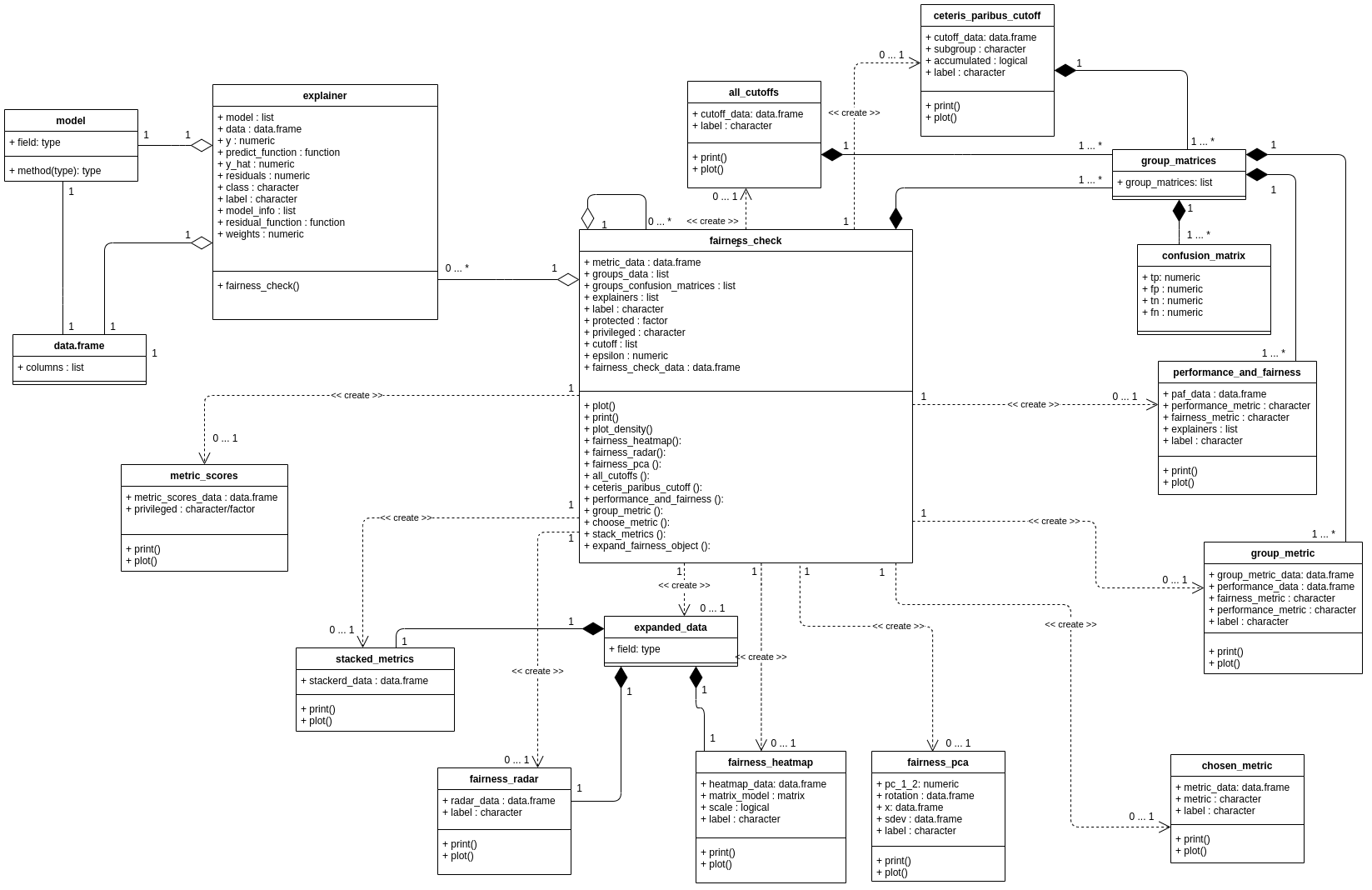}
  \caption{Class diagram for objects created by functions from the \fairmodels package. The bigger version of the plot can be found   \href{https://github.com/ModelOriented/fairmodels/blob/master/man/figures/class_diagram.png}{here}}
  \label{figure:classdiagram}
 \end{figure}
\end{landscape}

To create \texttt{fairness\_object}, at least one explainer needs to be passed to \texttt{fairness\_check()} function which returns the said object. While creating \textit{\fobject} metrics for all subgroups are calculated from confusion matrices. The \fobject has numerous fields, some of them are: 
\begin{itemize}
    \item \textbf{parity\_loss\_metric\_data} - data.frame containing parity loss for each metric and classifier,
    \item \textbf{groups\_data} - list of metric scores for each metric and model,
    \item \textbf{group\_confusion\_matrices} - list of values in confusion matrices for each model and metric,
    \item \textbf{explainers} - list of DALEX explainers. When explainers and/or \textit{\fobject} are added, then explainers and/or explainers extracted from \textit{\fobject} are added to that list,
    \item \textbf{label} - character vector of labels for each explainer.
    \item ... - other fields.
\end{itemize}
The \textit{\fobject} methods are used to create numerous objects that help to visualize bias. In next sections we list more detailed functions for deeper exploration of bias. Detailed relations between objects created with \fairmodels are depicted in Figure \ref{figure:classdiagram}. 
The general overview of the workflow is presented in Figure \ref{fig:flowchart}.

\section{Visualizing bias}
\label{visualization}

In \fairmodels there are 12 metrics based on confusion matrices for each subgroup, see Table \ref{tab:Metrics-table} for the complete list. Some of them were already introduced before. 
\begin{table}[h!]
\begin{center}
\footnotesize
    \begin{tabular}{p{1cm}p{2.6cm}p{3.8cm}p{4cm}}
    \hline
Metric & Formula & Name & Fairness criteria \\ \hline    
TPR & $\frac{TP}{TP + FN}$ & True positive rate & Equal opportunity \newline\citep{NIPS2016_6374} \\ \hline
TNR & $\frac{TN}{TN + FP}$ & True negative rate & \\ \hline
PPV & $\frac{TP}{TP + FP}$ & Positive predictive value & Predictive parity \newline\citep{ppv}\\ \hline
NPV & $\frac{TN}{TN + FN}$ & Negative predictive value & \\ \hline
FNR & $\frac{FN}{FN + TP}$ & False negative rate & \\ \hline
FPR & $\frac{FP}{FP + TN}$ & False positive rate & Predictive equality \newline\citep{ppe} \\ \hline
FDR & $\frac{FP}{FP + TP}$ & False discovery rate &\\ \hline
FOR & $\frac{FN}{FN + TN}$ & False omission rate & \\ \hline
TS & $\frac{TP}{TP + FN + FP}$  & Threat score &\\ \hline
STP & $\frac{TP + FP}{TP + FP + TN + FN}$ & Positive rate & Statistical parity \newline \citep{statisticalparity}\\ \hline
ACC & $\frac{TP + TN}{TP + TN + FP + FN}$ & Accuracy & Overall accuracy equality \newline\citep{accuracy} \\ \hline
F1 &  $\frac{2 \cdot PPV * TPR}{PPV + TPR}$ & F1 score &\\ \hline
    \end{tabular}
    \caption{Fairness metrics implemented in the \textbf{fairmodels} package}
\label{tab:Metrics-table}
\end{center}
\end{table}

Not all metrics are needed to determine if the discrimination exists but they are helpful to acquire a fuller picture. To facilitate the visualization over many subgroups, we introduce a function that maps metric scores among subgroups to a single value. This function, which we called \textit{parity\_loss}, has an attractive property. Due to the usage of the absolute value of the natural logarithm, it will return the same value whether the ratio is inverted or not.

So for example when we would like to know the parity loss of Statistical Parity between unprivileged (b) and privileged (a) subgroups we mean value like this: 
\begin{equation}
STP_{\parityloss} = \Big | \ln \Big( \frac{STP_b}{STP_a} \Big)\Big|. 
\end{equation}
This notation is very helpful because it allows to accumulate $STP_{\parityloss}$ overall unprivileged subgroups, so not only in the binary case
\begin{equation}
STP_{\parityloss} = \sum_{i \in \{a, b, ...\}} \Big|\ln \Big(\frac{STP_i}{STP_a} \Big)\Big|.  \label{eq:parityLoss}
\end{equation}

The \parityloss relates strictly to ratios. The classifier is more fair if \parityloss is low. This property is helpful in visualizations.

There are several modifying functions that operate on \texttt{fairness\_object()}. Their usage will return other objects which will be visualized in the following chapter on the class diagram (Fig \ref{figure:classdiagram}). The objects can then be plotted with generic \code{plot()} function. Additionally, there is a special plotting function that works immediately on \fobject which is \code{plot\_density}. In some functions, the user can directly specify which metrics shall be visible in the plot. The detailed technical introduction for all these functions is presented in \href{https://cran.r-project.org/web/packages/fairmodels/fairmodels.pdf}{\fairmodels manual}.
Plots visualizing different aspects of \parityloss can be created with one of following pipelines:
\begin{itemize}
    \item \textbf{ \fobject \%$>$\% modifying\_function(...) \%$>$\% plot()}
    
    This pipe is preferred and allows setting parameters in both modifying functions and certain plot functions which is not the case with the next pipeline. 
    \item \textbf{ \fobject \%$>$\% plot\_fairmodels(type = modifying\_function, ...)}
    
    Additional parameters are passed to the modifying functions and not to the plot function.
\end{itemize}

\begin{figure}[ht]
\centering

\subfloat[metric\_scores]{
  \includegraphics[width=0.45\textwidth]{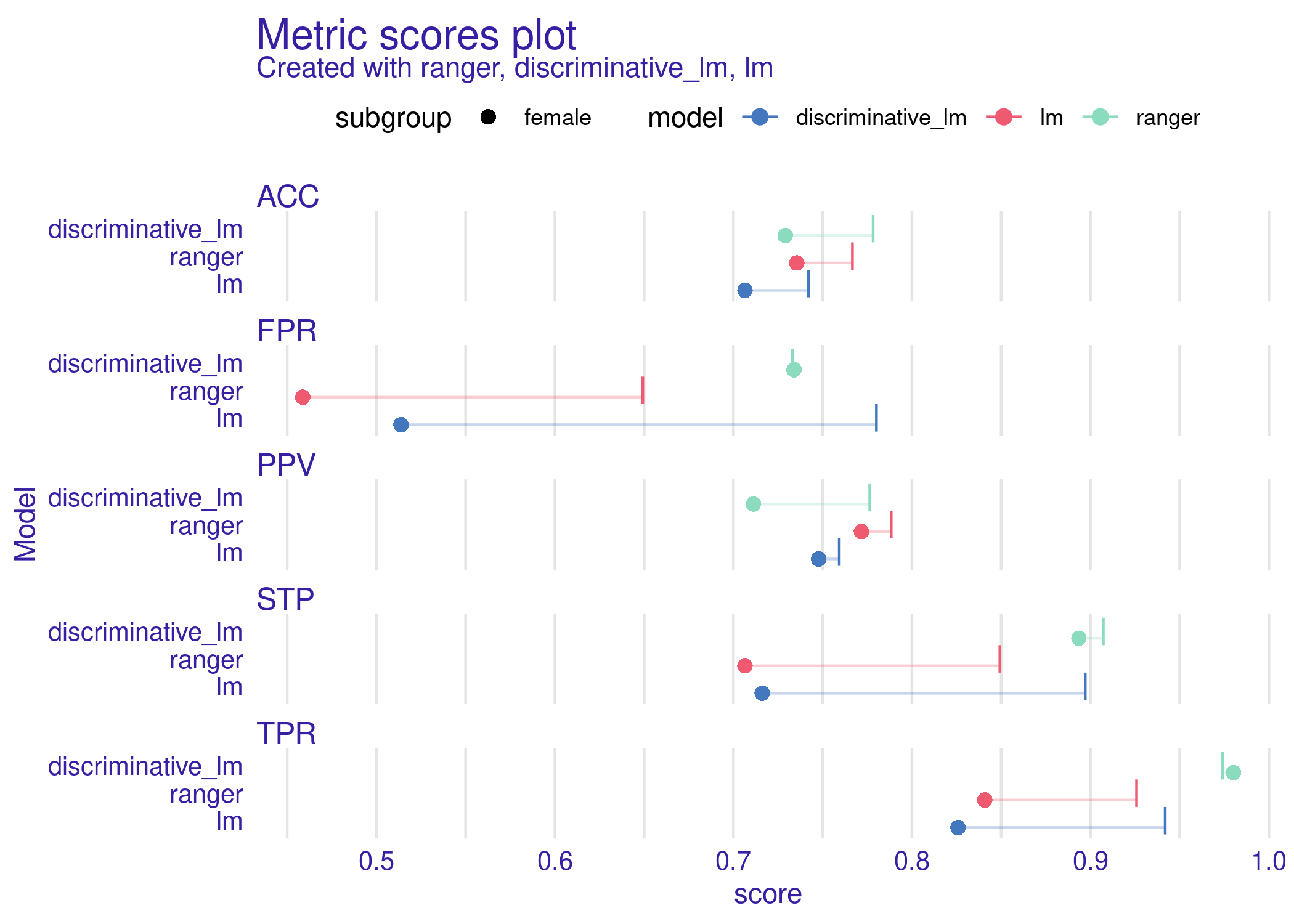}
}
\subfloat[fairness\_radar]{
  \includegraphics[width=0.45\textwidth]{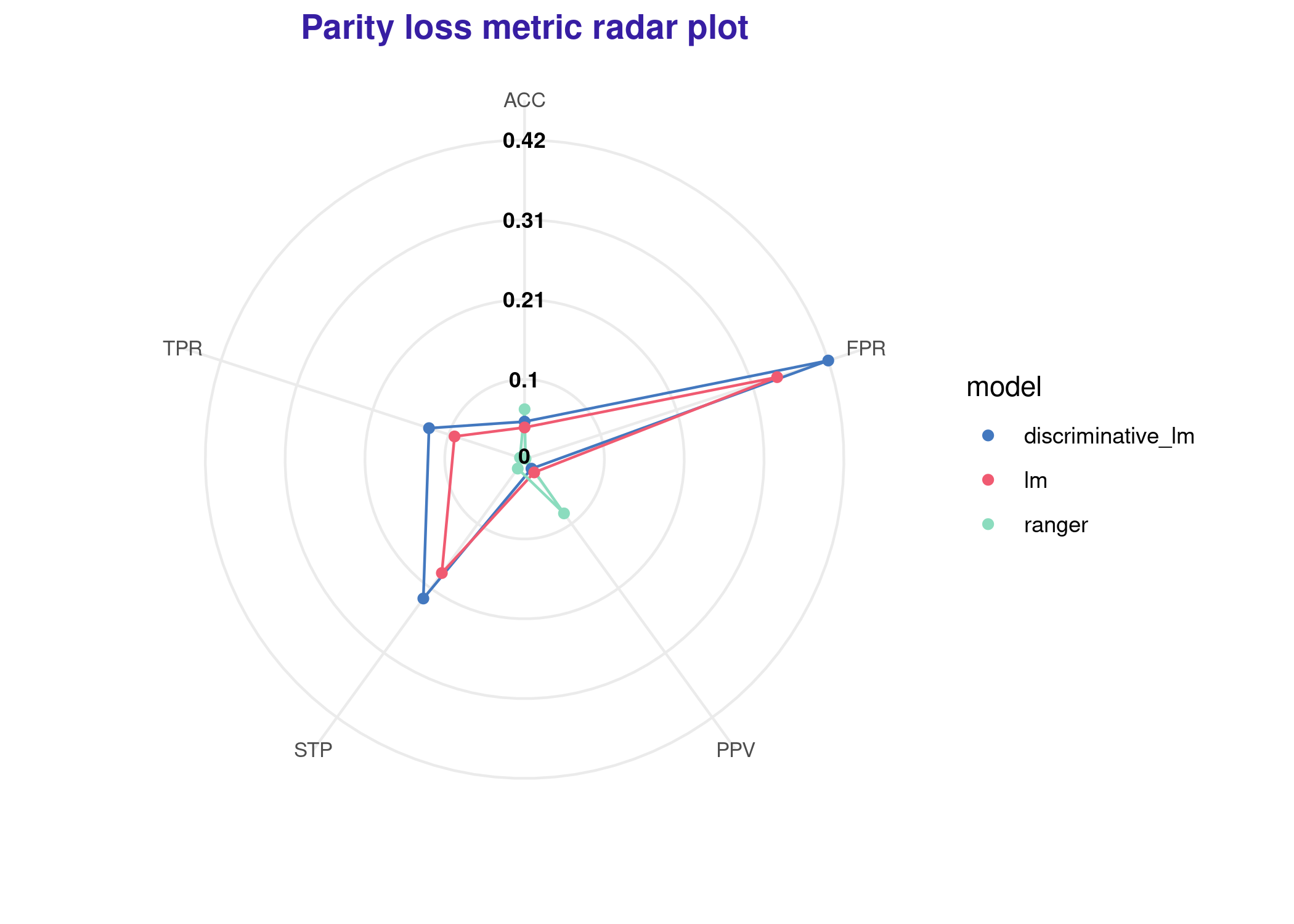}
}

\hspace{0mm}

\subfloat[fairness\_heatmap]{
  \includegraphics[width=0.47\textwidth]{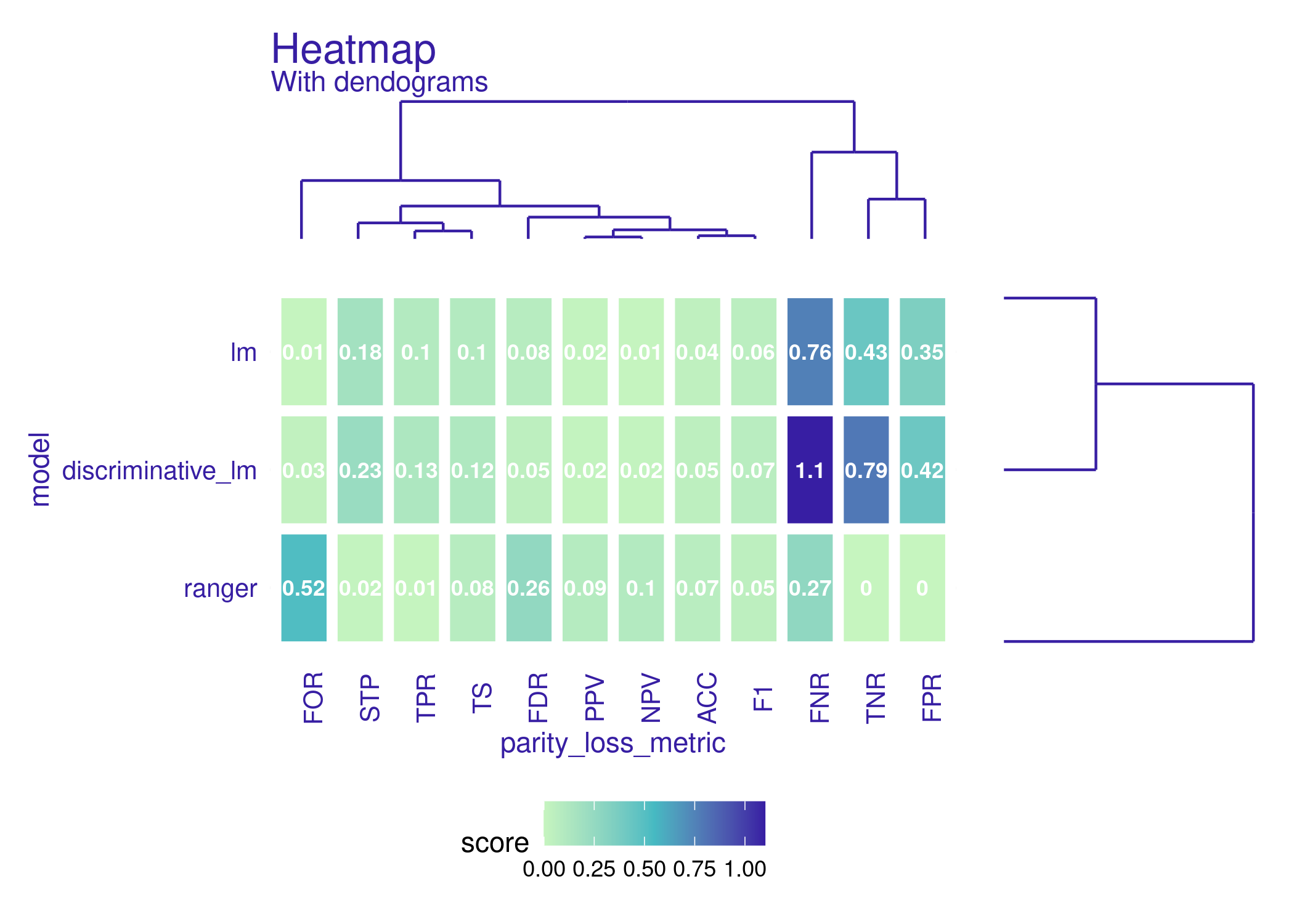}
}
\subfloat[fairness\_pca]{
  \includegraphics[width=0.47\textwidth]{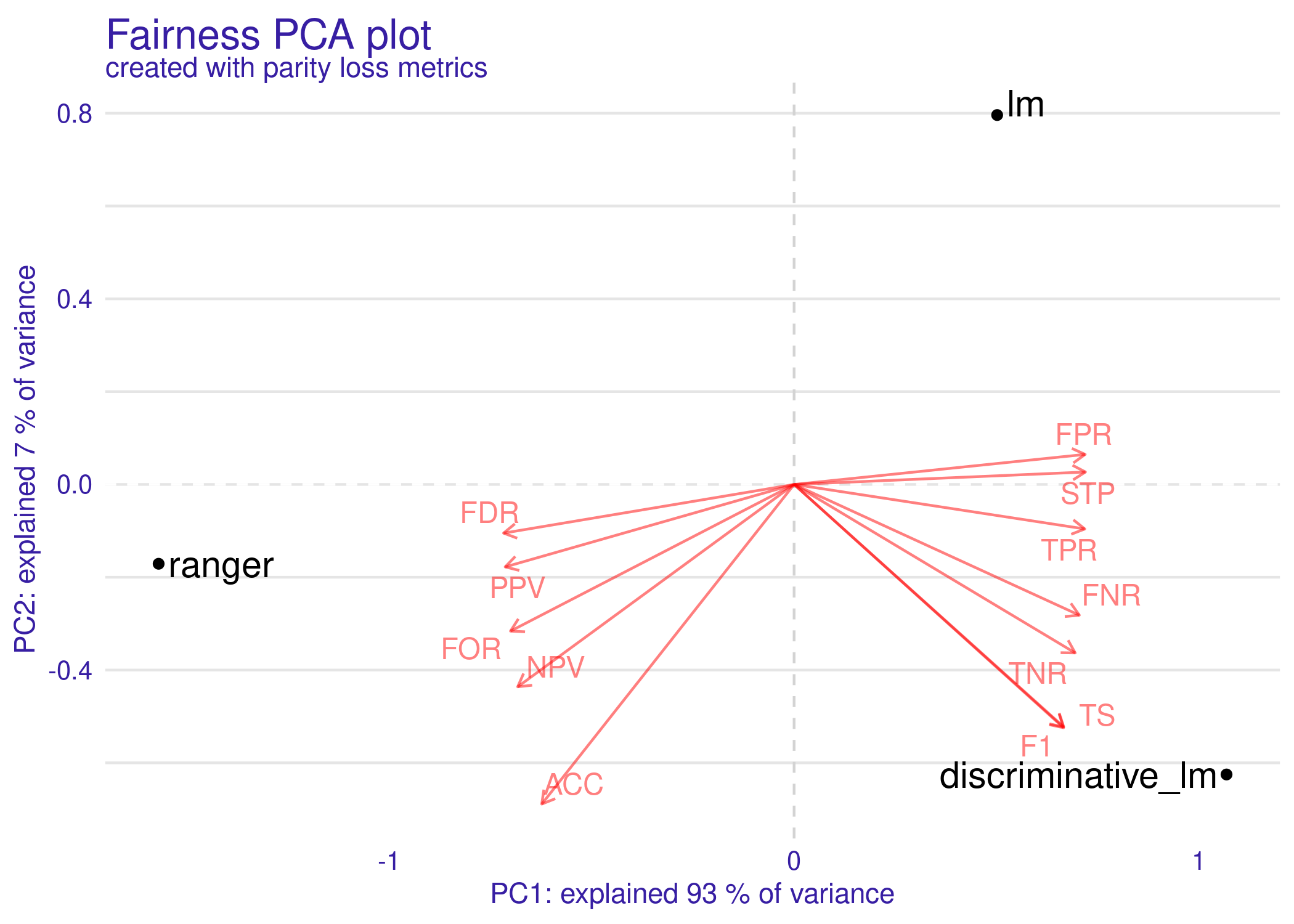}
}

\hspace{0mm}

\subfloat[choose\_metric]{
  \includegraphics[width=0.47\textwidth]{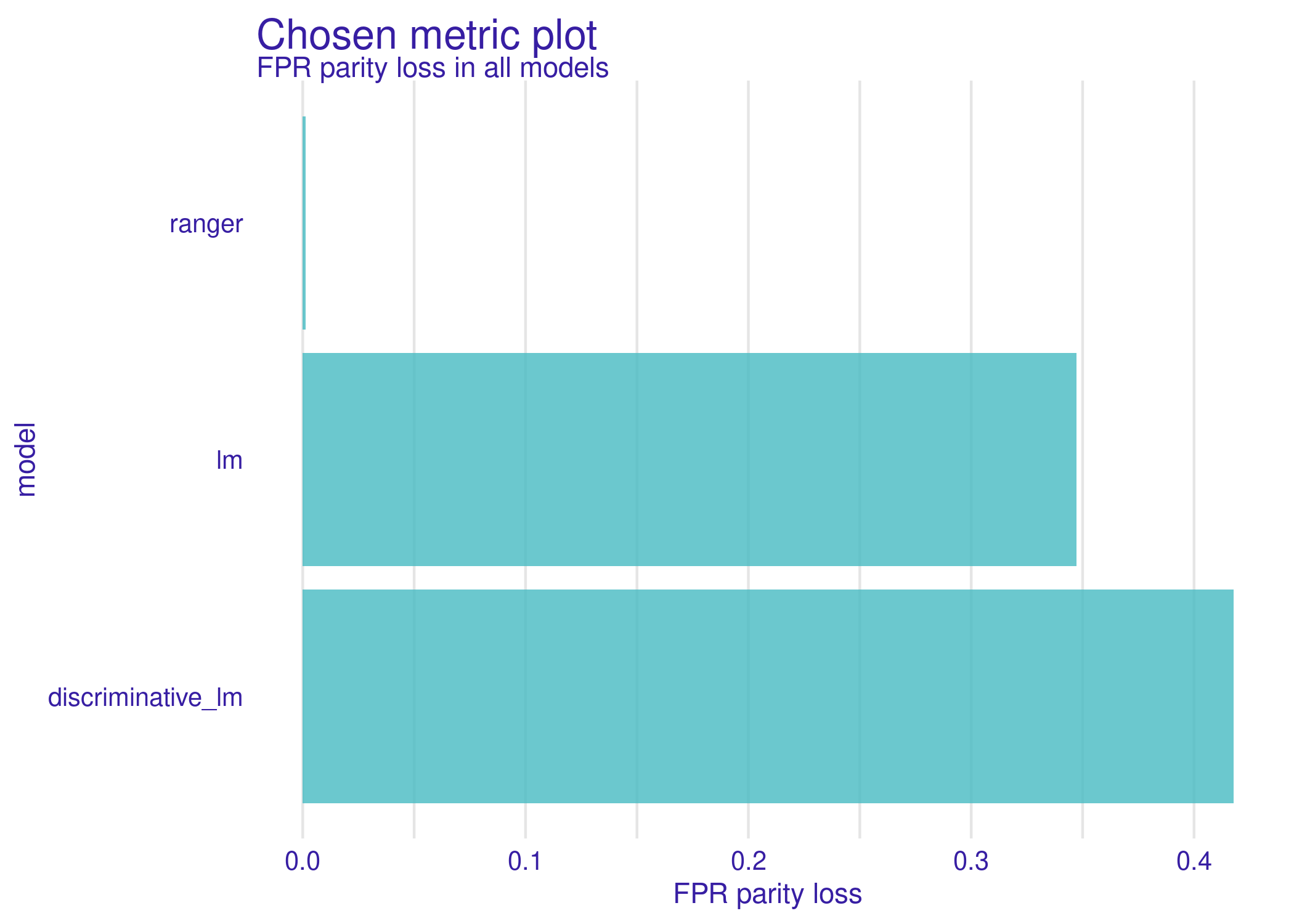}
}
\subfloat[stack\_metrics]{
  \includegraphics[width=0.47\textwidth]{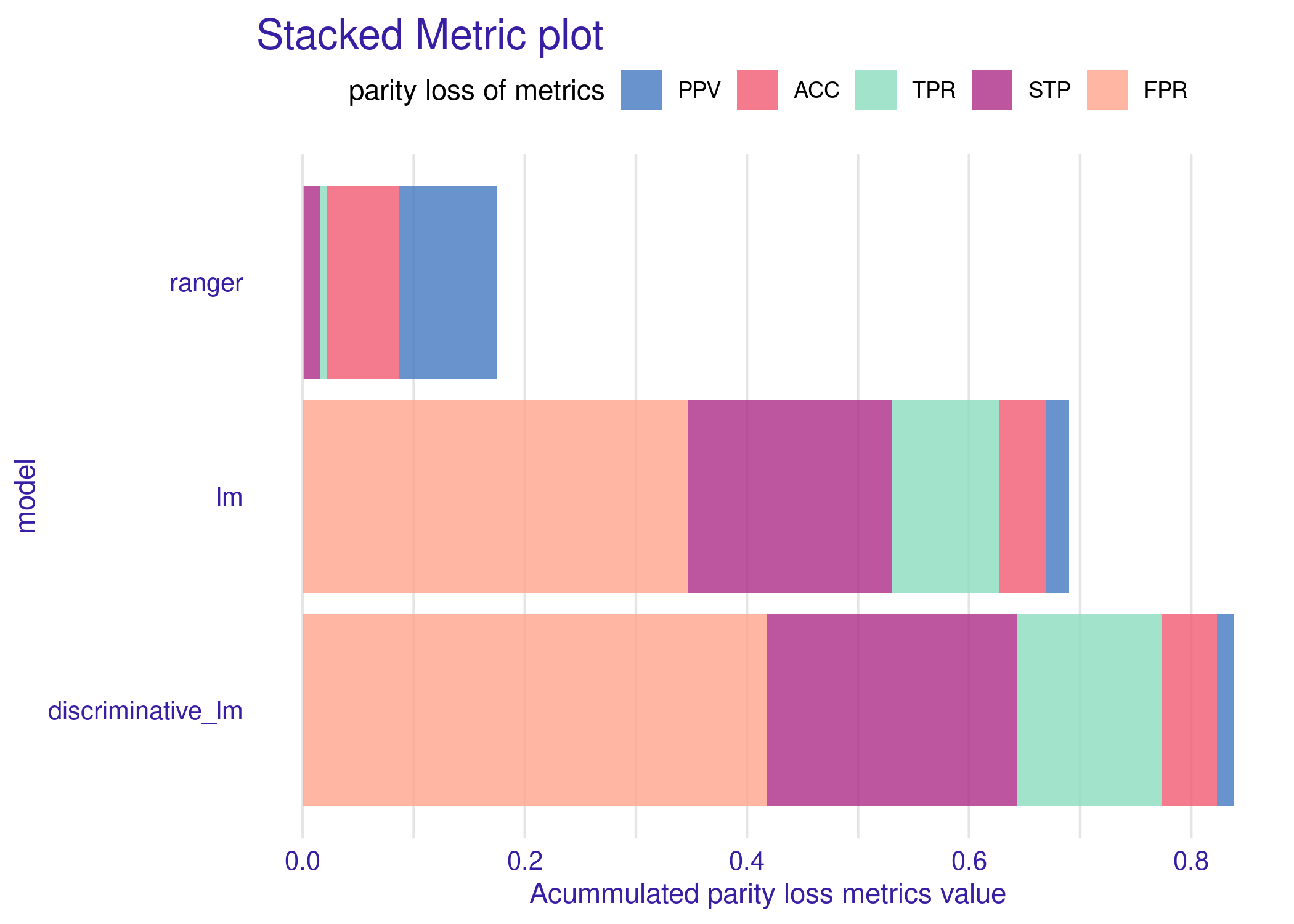}
}
\hspace{0mm}

\subfloat[group\_metric]{
  \includegraphics[width=0.47\textwidth]{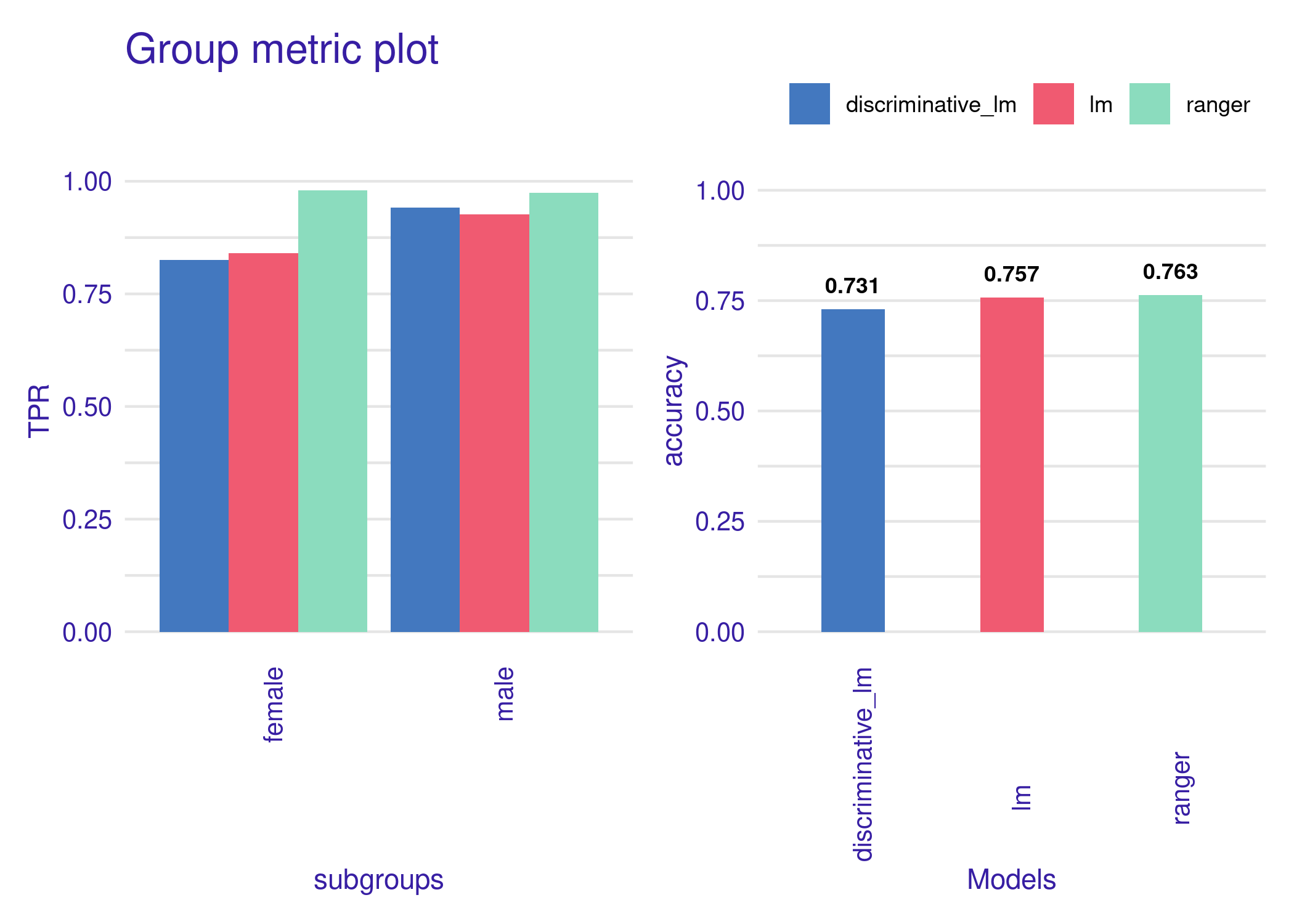}
}
\subfloat[plot\_density]{
  \includegraphics[width=0.47\textwidth]{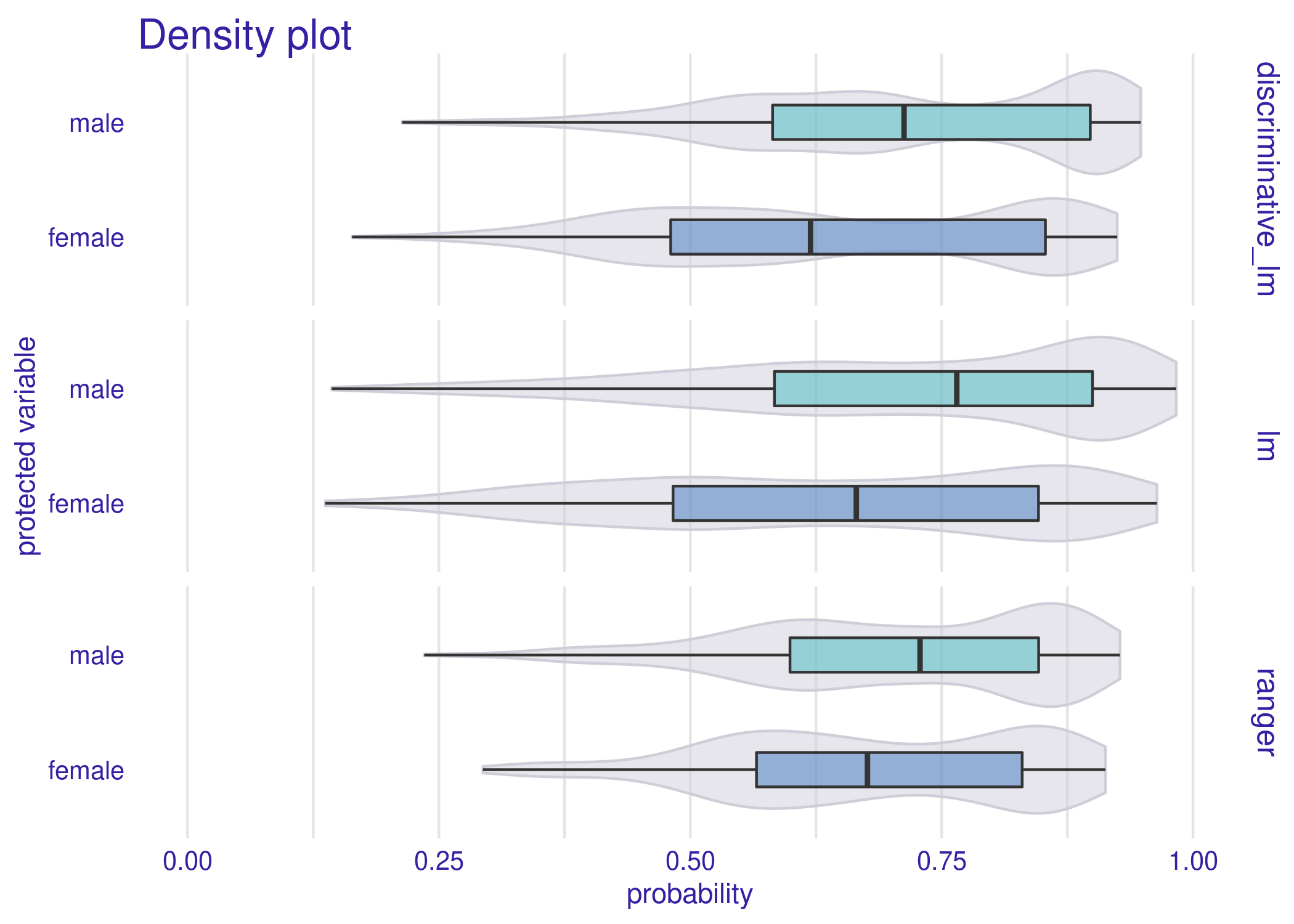}
}
\hspace{0mm}

\end{figure}
\clearpage

\begin{figure}[ht]
\centering

\renewcommand{\thesubfigure}{i}
\subfloat[performance\_and\_fairness]{
    
  \includegraphics[width=0.47\textwidth]{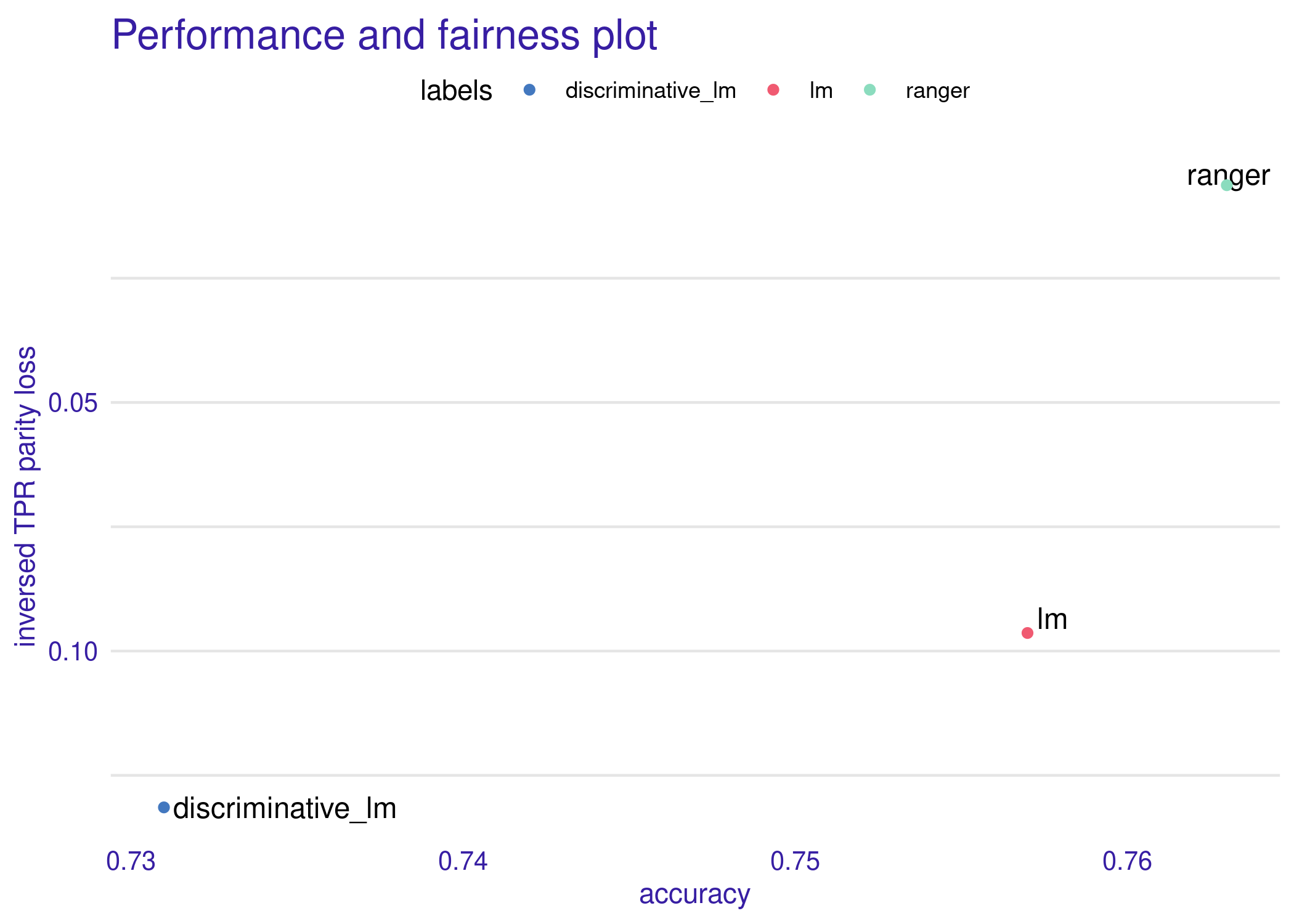}
}
\renewcommand{\thesubfigure}{j}
\subfloat[all\_cutoffs]{
  \includegraphics[width=0.47\textwidth]{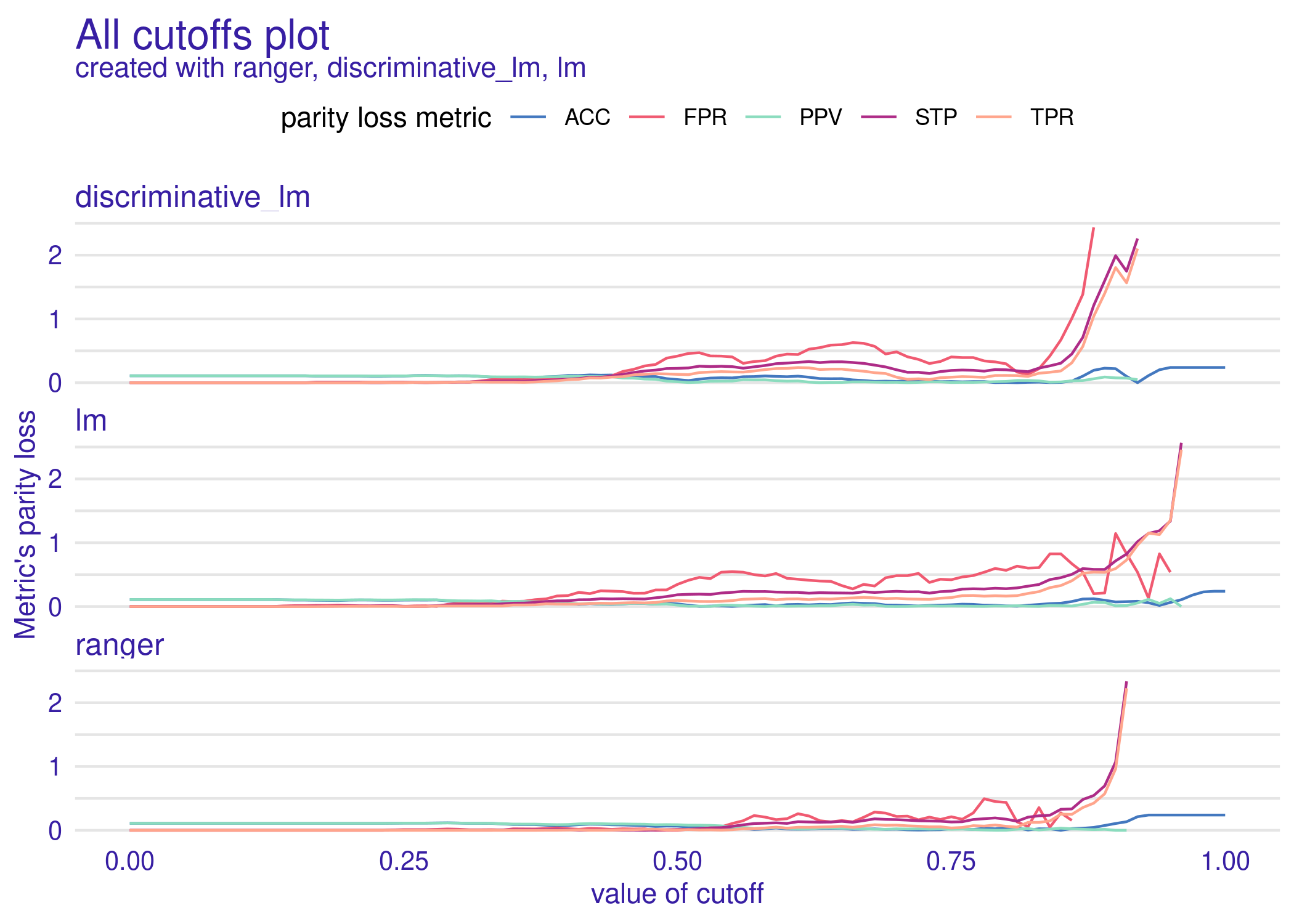}
}
\hspace{0mm}

\renewcommand{\thesubfigure}{k}
\subfloat[ceteris\_paribus\_cutoff]{
  \includegraphics[width=0.47\textwidth]{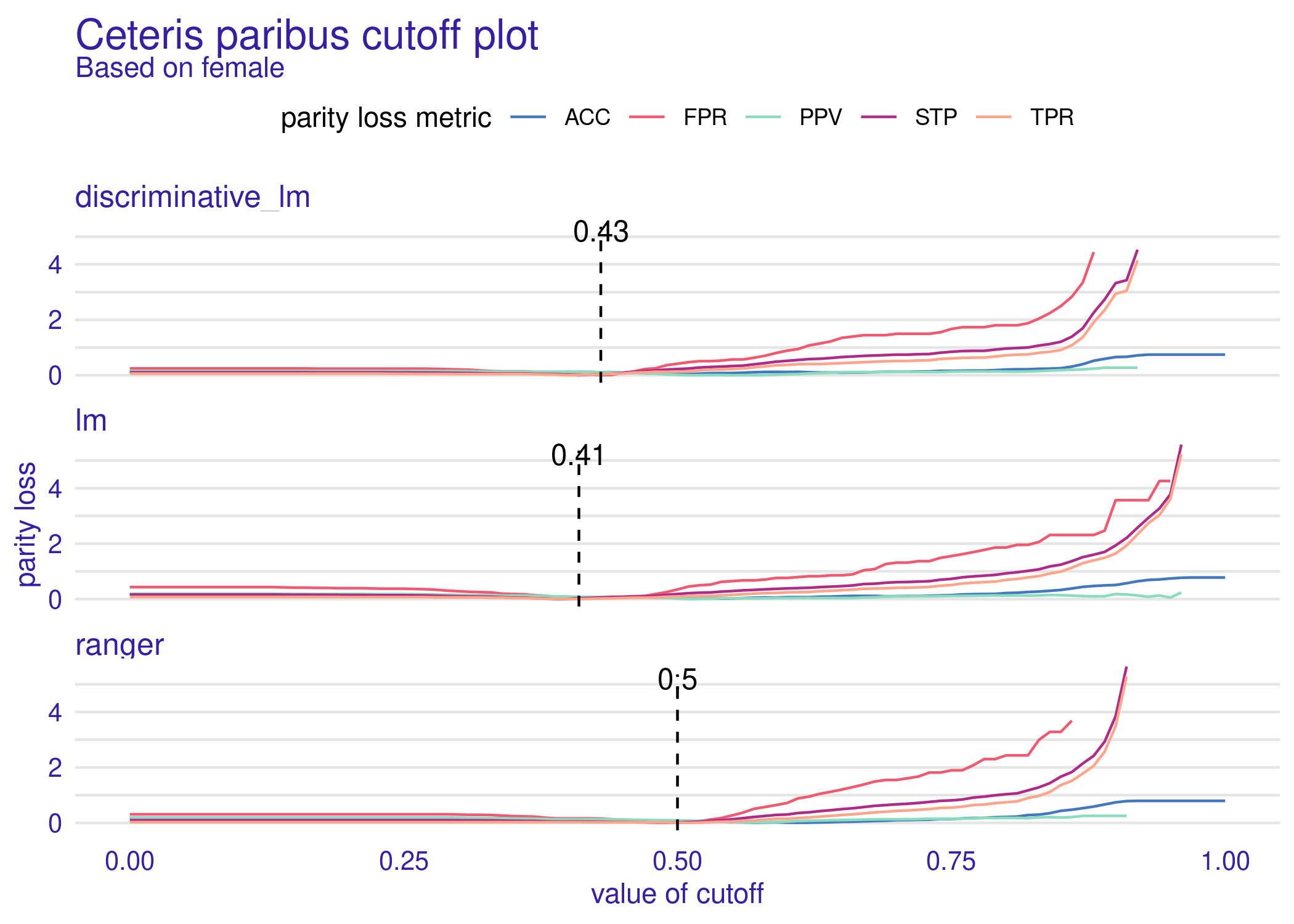}
}
\renewcommand{\thesubfigure}{l}
\subfloat[ceteris\_paribus\_cutoff with parameter cumulated = TRUE]{
  \includegraphics[width=0.47\textwidth]{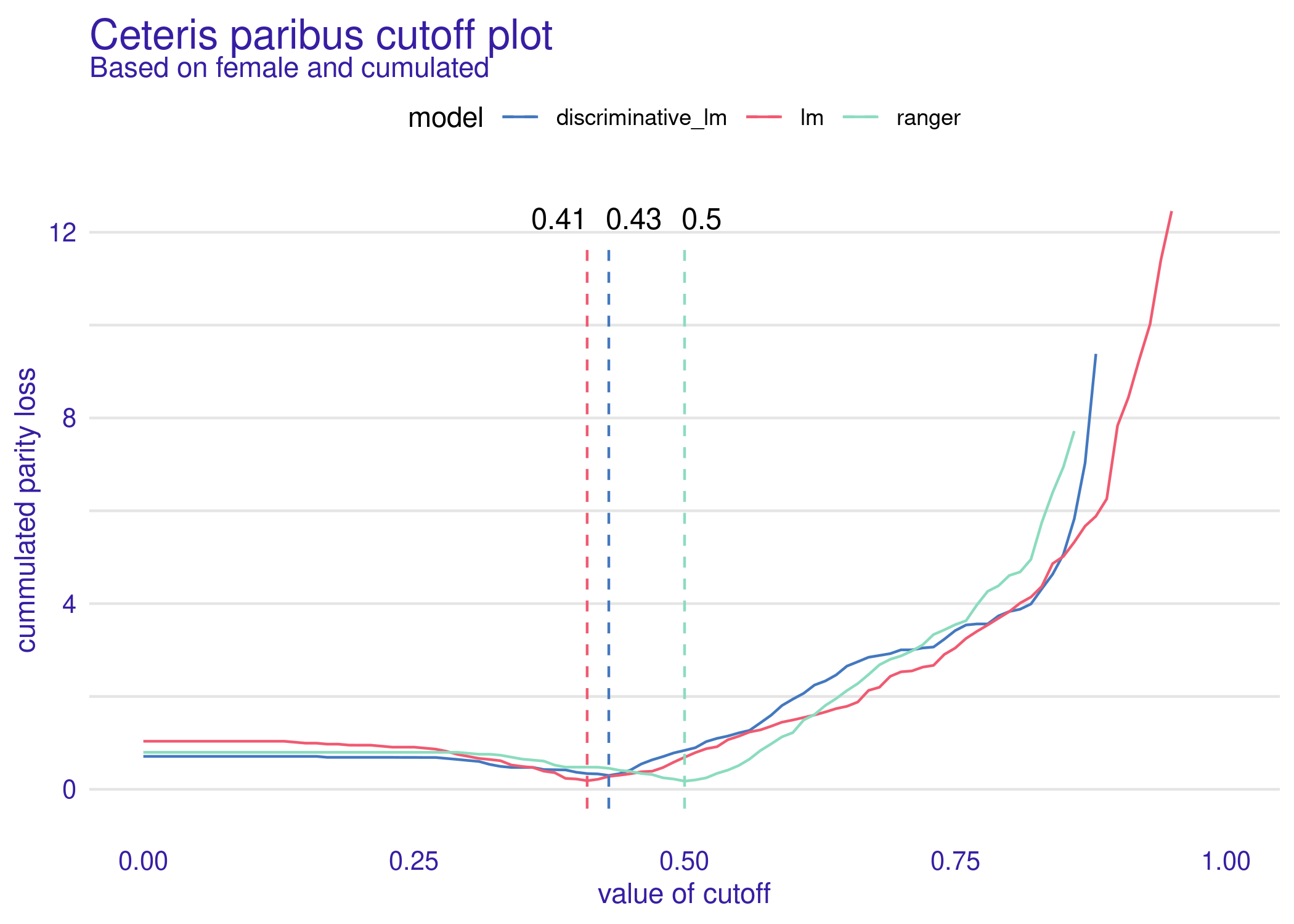}
}
\hspace{0mm}
\caption{Figure depicts twelve methods for a more detailed exploration of biases. The subtitle of each graph indicates the name of the function that produces that type of graph.}
\label{fig:all}
\end{figure}

By using the pipelines, different kinds of plots can be obtained by superseding the \textbf{modifying\_function} with function names (a-l). They can be seen in Figure \ref{fig:all}. To see different aspects of fairness and bias, a user can choose the model with the smallest bias (b, e, f), find out the similarity between metrics and between models (c, d), compare models in both fairness and performance (i, g), and see how cutoff manipulation might change the \parityloss (j, k, l). 

Apart from methods (b-l), there is one special one, more related to plotting \code{fairness\_check} (Figure \ref{fig:fc1} and \ref{fig:fc2_plot}), but uses its product - \fobject like other visualization methods - function called \code{metric\_scores} (a) shows exact rates of metrics for each model and subgroup. 

\section{Bias mitigation}
\label{mitigation}
What can be done if the model does not meet the fairness criteria? Machine learning practitioners might try to use other algorithms or sets of variables to construct unbiased models, but this does not guarantee to pass the \texttt{fairness\_check()}. An alternative is to use bias mitigation techniques that adjust the data or model so that fairness conditions are met.
There are essentially three types of such methods. The first is data pre-processing. When there are unwanted correlations between variables or sample sizes among subgroups in data, there are many ways to "correct" the data. The second one is in-processing, which is for example optimizing classifiers not only to reduce classification error but also to minimize a fairness metric. Last but not least is post-processing which modifies model output so that predictions and miss-predictions among subgroups are more alike. 

The \fairmodels package offers five functions for bias mitigation, three for pre-processing, and two for post-processing algorithms. Most of these approaches are also implemented in  \href{https://aif360.mybluemix.net/}{aif360} \citep{aif360-oct-2018} although in \fairmodels there are separate implementations of them in R. There are a lot of useful mitigation techniques that are not in \fairmodels like \cite{NIPS2016_6374} and numerous in-processing algorithms.

\subsection{Data Pre-processing}
\begin{itemize}
\item \textbf{Disparate impact remover}
    
    In \fairmodels geometric repair, an algorithm originally introduced by \cite{disparateImpact}, works on ordinal, numeric features. Depending on the $\lambda \in [0,1]$ parameter, this method will transform the distribution of a given feature. The idea is simple, given feature distribution in different subgroups the algorithm finds optimal distribution (according to earth mover's distance) and transforms distribution for each subgroup to match the optimal one. For example, if the age is an important feature and its distribution is different in two subgroups, and we want to change that then the geometric repair will map each individual's age to a new distribution (different age). It will be preserving the order - the ranks (in our case seniority) of observations are preserved. Parameter $\lambda$ is responsible for the repair degree, so for full repair lambda should be set to 1. The method does not focus on a particular metric, but rather tries to level out the level playing field by transforming potentially harmful feature distributions.
\item \textbf{Reweighting}

Reweighting is a rather simple approach. This method was implemented according to \cite{kamiran}. It computes weights by dividing the theoretical probability of assigning favorable label for a subgroup by real (observed) probability (based on the data). Theoretic probability for a subgroup is computed by multiplying the probability of assigning favorable label (for all populations) by the probability of picking observation from a certain subgroup. It focuses on mitigating statistical parity.
\item \textbf{Resampling}

Resampling bases on weights calculated in \texttt{reweighting}. Each weight for a subgroup is multiplied by the size of the subgroup. Then, whether the subgroup is deprived or not (if weight is higher than one the subgroup is considered deprived), observations are duplicated from either ones that were assigned a favorable label or not. There are two types of resampling- uniform and preferential. The uniform is making algorithm pick or omit observations randomly without taking into consideration its probabilistic score. Preferential is making use of another probabilistic classifier, potentially different from the main model for final predictions. In \cite{kamiran} it is called ranker - it predicts the probabilities for the observations to decide which observations are close to the cutoff border (usually 0.5). Based on the probabilistic output of the ranker, the observations are sorted and the ones with the highest/lowest ranks are either left out or duplicated depending on the case. More on that on \cite{kamiran}. The \fairmodels implementation instead of training the ranker as in the aforementioned paper uses a vector of previously calculated probabilities provided by the user. With this, it shifts the decision and responsibility of choosing a ranker to the user. It focuses on mitigating statistical parity.  
\end{itemize}

\subsection{Model Post-processing}
\begin{itemize}
\item \textbf{Reject Option based Classification Pivot}

Based on \cite{postKamiran} in \fairmodels \textit{roc\_pivot} method was implemented. Let $\theta \in (0,1)$ be the value that determines the radius of the so-called critical region, which is an area around the cutoff. The $\theta$ is specified by the user and it should describe how big the critical region should be. For example if $\theta = 0.1$ and cutoff is 0.6, then the critical region will be (0.5, 0.7). Let's assume that we are predicting a favorable outcome. If the assigned probability of observation is in the described region, then with a certain assumption the probabilities are pivoting on the other side of the cutoff. If an observation in a critical region is considered to be the privileged and it is on the right side of the cutoff, then its probabilities are pivoting from the right side of the cutoff to the left. So if an observation is in the critical region and it is considered unprivileged, then if it is on the left side of the cutoff it will pivot to the right side. Pivoting here means changing the side of the cutoff so that the distance from the cutoff stays unchanged. It does not intend to mitigate a single metric but rather changes predictions in the critical region that are not as certain as others which might lower more metrics.  

\item \textbf{Cutoff manipulation}

The \fairmodels package supports setting cutoff for each subgroup. User may pick \parityloss metrics of their choice and find the minimal \textit{parity\_loss}. It is part of \texttt{ceteris\_paribus\_cutoff()} function. Based on picked metrics, the sum of parity loss is calculated for each cutoff of the chosen subgroup. Then the minimal value is found. This way optimal values might be found for metrics of interest. The minimum is marked with a dashed vertical line (see Figure \ref{fig:all} subplot l). This approach however might be to some extent concerning. Some might argue that this method of setting different cutoffs for different subgroups is unfair and is punishing privileged subgroups for something that they have no control of. Especially in the individual fairness field, it would be concerning if 2 similar people with different sensitive attributes would have 2 different thresholds and potentially 2 different outcomes.  This is a valid point and this method should be used with knowledge of all its drawbacks. The cutoff manipulation method targets metrics chosen by the user.
\end{itemize}

Methods listed above similarly to visualizing have two possible pipelines. All pre-processing methods can be used with 2 pipelines whereas post-processing can be used in one specific way.
\begin{enumerate}
    \item Pre-processing pipelines
\begin{itemize}
    \item \textbf{data/explainer \%>\% method }
        
Returns either weights, indexes, or changed data depending on the method used.
    \item \textbf{data/explainer \%>\% pre\_process\_data(data, protected, y, type = ...)}

Always returns data.frame. In case of weights data has additional column called \verb'_weights_'. 
\end{itemize}
\item Post-processing pipelines
\begin{itemize}
    \item \textbf{\fobject \%>\% ceteris\_paribus\_cutoff(subgroup, ...) \%>\% print()/plot()}

This is the pipeline for creating ceteris paribus cutoff print and plot.
\item \textbf{explainer \%>\% roc\_pivot(protected, privileged, ...)}

The pipeline will return explainer with \verb'y_hat' field changed.
\end{itemize}
\end{enumerate}

The User should be aware that debiasing one metric might enhance bias in another. It is a so-called fairness-fairness trade-off. There is also a fairness-performance trade-off where debiasing one metric leads to worse performance. Another thing to remember is as found in \cite{Agrawal2020DebiasingCI} metrics might not generalize well to out-of-distribution examples, so it is advised to also check the fairness metrics on a separate test set.

\subsection{Example}

Now we will show an example usage of one pre-processing and one post-processing method. As before, the   \textbf{German Credit Data} will be used along with previously created \code{lm\_model}. Firstly, we create a new dataset using \code{pre\_process\_data} and then we use it to train the logistic regression classifier.

\begin{example}

resampled_german   <- german 
                                                  y_numeric,
                                                  type = 'resample_uniform')

lm_model_resample  <- glm(Risk~.,
                         data   = resampled_german,
                         family = binomial(link = "logit"))

explainer_lm_resample <- DALEX::explain(lm_model_resample,
                                        data = german[,-1],
                                        y = y_numeric)
\end{example}
Then we make other explainer. We use previously created \code{explainer\_lm} with post-processing function \code{roc\_pivot}. We set parameter \code{theta = 0.05} for rather narrow area of pivot.

\begin{example}
new_explainer <- explainer_lm 
                                            privileged = "male", 
                                            theta = 0.05)
\end{example}
In the end, we create \fobject with explainers obtained with code above and one created in the first example to see the difference.  
\begin{example}

fobject <- fairness_check(explainer_lm_resample, new_explainer, explainer_lm,
                          protected = german$Sex,
                          privileged = "male",
                          label = c("resample", "roc", "base"))

fobject 

\end{example}

\begin{figure}
\includegraphics[width=0.9\linewidth]{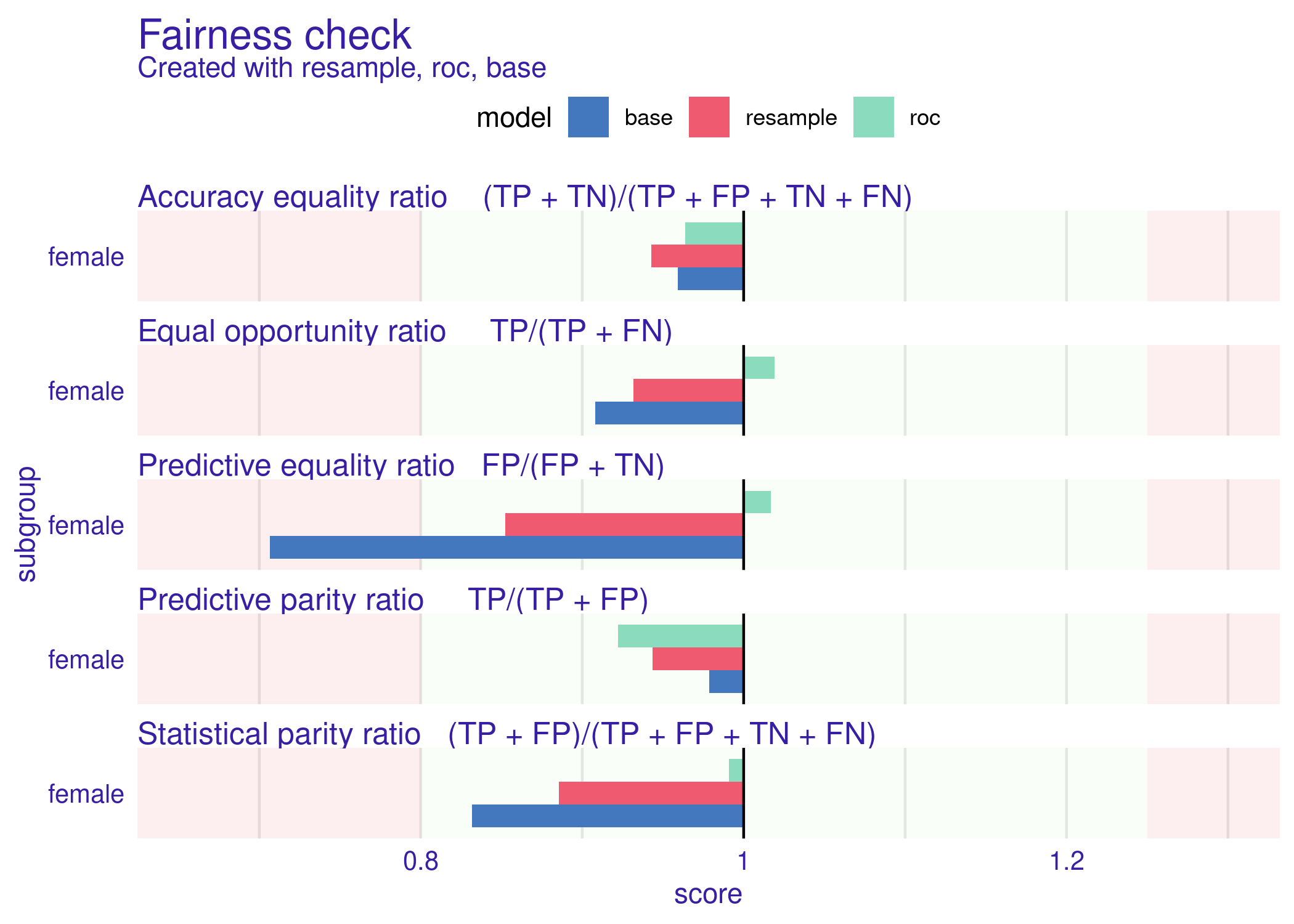} 
\caption{Graphical summary of a base model and model after two bias mitigation techniques. }
\label{fig:mitigation}
\end{figure}

The result of the code above is presented in Figure \ref{fig:mitigation}.  The mitigation methods successfully eliminated bias in all of the metrics. Both models are better than the original \code{base}. This is not always the case - sometimes eliminating bias in one metric may increase bias in another metric. For example, let's consider a model which is perfectly accurate, but some subgroups receive few positive predictions (bias in Statistical parity). In that case, mitigating the bias in Statistical parity would decrease accuracy.

\section{Summary and future work}
In this paper, we showed that checking for bias in machine learning models can be done conveniently and flexibly. The package \fairmodels described above is a self-sufficient tool for bias detection, visualization, and mitigation in classification machine learning models. We presented theory, package architecture, suggested usage of package, and examples along with plots. Along the way, we introduced the core concepts and assumptions that come along the bias detection and plot interpretation. The package is still improved and enhanced which can be seen by adding the announced regression module based on \cite{regression}. We did not cover it in this article because it is still an experimental tool. Another tool for in-processing classification closely related to \fairmodels has also been added and can be found on \href{https://github.com/ModelOriented/FairPAN}{https://github.com/ModelOriented/FairPAN}.

The source code of the package, vignettes, examples, and documentation can be found at \newline \href{https://modeloriented.github.io/fairmodels/}{https://modeloriented.github.io/fairmodels/}. The stable version is available on CRAN. The code and the development version can be found on GitHub \href{https://github.com/ModelOriented/fairmodels}{https://github.com/ModelOriented/fairmodels}. This is also a place to report bugs or requests (through GitHub issues).

In the future, we plan to enhance the spectrum of bias visualization plots and introduce methods for regression and individual fairness. The potential way to explore would be an in-processing bias mitigation - training models that minimize cost function and adhere to certain fairness criteria. This field is heavily developed in Python and lacks appropriate attention in R.  

\section*{Acknowledgements}
Work on this package was financially supported by the NCN Sonata Bis-9 grant 2019/34/E/ST6/00052.

\bibliography{wisniewski-biecek}

\address{Jakub Wiśniewski\\
  Faculty of Mathematics and Information Science\\
  Warsaw University of Technology\\
  Poland\\
  \email{jakwisn@gmail.com}}

\address{Przemysław Biecek\\
  Faculty of Mathematics and Information Science\\
  Warsaw University of Technology\\
  Faculty of Mathematics, Informatics, and Mechanics\\
  University of Warsaw\\
  Poland\\
  ORCiD: 0000-0001-8423-1823\\
  \email{przemyslaw.biecek@pw.edu.pl}}

\end{article}

\end{document}